\documentclass[10pt,twocolumn]{article}
\usepackage{graphicx}
\usepackage{amsmath,amssymb,times,cite}
\usepackage{xcolor}
\usepackage{multirow}
\usepackage{verbatim}

\newcommand{\op}[1]{\operatorname{#1}}
\newcommand{\bg}[1]{\boldsymbol{#1}} 
\newcommand{\bm}[1]{\mathbf{#1}} 

\newcommand\T{{\mathpalette\raiseT\intercal}}
\newcommand\raiseT[2]{%
\setbox0\hbox{$#1{#2}$}\raise\dp0\box0}

\usepackage{booktabs}
\usepackage[extra]{tipa}
\usepackage{accents}

\newlength{\dhatheight}

\title{\LARGE\textbf{Classification of Developmental and Brain Disorders via Graph Convolutional Aggregation}}
\author{
Ibrahim Salim and A. Ben Hamza\\
Concordia Institute for Information Systems Engineering\\
Concordia University, Montreal, QC, Canada
}
\date{}

\begin{document}
\maketitle

\begin{abstract}
While graph convolution based methods have become the de-facto standard for graph representation learning, their applications to disease prediction tasks remain quite limited, particularly in the classification of neurodevelopmental and neurodegenerative brain disorders. In this paper, we introduce an aggregator normalization graph convolutional network by leveraging aggregation in graph sampling, as well as skip connections and identity mapping. The proposed model learns discriminative graph node representations by incorporating both imaging and non-imaging features into the graph nodes and edges, respectively, with the aim of augmenting predictive capabilities and providing a holistic perspective on the underlying mechanisms of brain disorders. Skip connections enable the direct flow of information from the input features to later layers of the network, while identity mapping helps maintain the structural information of the graph during feature learning. We benchmark our model against several recent baseline methods on two large datasets, Autism Brain Imaging Data Exchange (ABIDE) and Alzheimer's Disease Neuroimaging Initiative (ADNI), for the prediction of autism spectrum disorder and Alzheimer's disease, respectively. Experimental results demonstrate the competitive performance of our approach in comparison with recent baselines in terms of several evaluation metrics, achieving relative improvements of 50\% and 13.56\% in classification accuracy over graph convolutional networks on ABIDE and ADNI, respectively.
\end{abstract}


\section{Introduction}
Understanding how the brain develops is vital to designing prediction models and formulating treatments for a variety of developmental disorders and degenerative neurological diseases such as autism spectrum disorder and Alzheimer's disease, which are devastating illnesses that have touched the lives of millions of families around the world, not only in personal anguish, but also in soaring healthcare costs~\cite{insel:15}. Autism spectrum disorder is a neurodevelopmental disability that affects how a person communicates, learns and socializes with others, whereas Alzheimer's disease is a chronic neurodegenerative brain disorder that slowly destroys brain cells, causing memory loss and cognitive decline over time.

Graph-structured data is prevalent across a diverse array of real-world application domains, including social networks, biological protein-protein interaction networks, molecular graph structures, and brain connectivity networks. Graphs offer a versatile means of representing real-world entities as a collection of nodes and their interactions as a series of edges. A case in point: for brain analysis in populations and diagnosis, we model populations as graphs, where each node represents a subject with an associated node feature vector obtained from imaging data, and each edge represents a pairwise similarity between two subjects with an edge feature vector acquired from non-imaging data.

In recent years, there has been a surge of interest in extending deep learning approaches to non-Euclidean domains thanks, in large part, to the prevalence and increasing proliferation of graph-structured data~\cite{kipf:17,wu:19,zeng:19,chen:20}. Advances in deep learning have
spawned significant efforts to facilitate, for instance, the clinical diagnosis of brain diseases. Graph convolutional networks (GCNs), which generalize convolutional neural networks to graph-structured data by leveraging spectral graph theory and its extensions, have gained popularity in graph representation learning~\cite{kipf:17} for their ability to capture the graph structure. GCN applies a layer-wise propagation rule that utilizes a first-order approximation of spectral graph convolutions. This involves updating the feature vector of each node in the graph by computing a weighted sum of the feature vectors of its neighboring nodes that are immediately connected to it. Wu \textit{et al.}~\cite{wu:19} propose a simple method for graph convolution, which involves eliminating the non-linear transition functions between the layers of a graph convolutional network. This results in collapsing the resulting function into a single linear transformation using the powers of the normalized adjacency matrix, with the addition of self-loops for all nodes in the graph. The simple graph convolution, however, functions as a low-pass filter that dampens all frequencies except for the zero frequency. This leads to oversmoothing. Zeng \textit{et al.}~\cite{zeng:19} introduce a graph sampling based learning method by sampling the training graph in lieu of nodes or edges across GCN layers, as well as eliminating biases in minibatch estimation via aggregator normalization techniques. Chen \textit{et al.}~\cite{chen:20} propose an extension to the GCN model that addresses the oversmoothing issue commonly observed with increasing network depth. This extension incorporates skip connections from the input layer and utilizes identity mapping along with a trainable weight matrix for each layer.

The primary objective of graph convolution based methods is to learn node representations that encode structural information about the graph. These learned node representations can then be used as input to machine learning models for downstream tasks such as node classification whose goal is to predict the most probable labels of nodes in a graph. For instance, in brain diagnosis tasks, which is the focus of this work, we want to classify subjects as diseased or healthy by predicting the node labels in a population graph. Graph convolution based methods have recently become prevalent in the biomedical and medical imaging domains~\cite{khosla:19,gopinath:19,su:20,yue:20,yang:20,Zhang22autism} due largely to the fact that neuroimaging provides valuable information about the diagnosis and progression of brain diseases. Built on top of graph signal processing approaches~\cite{goldsberry:17}, GCNs have shown promising results in metric learning and classification tasks on brain connectivity networks~\cite{ktena:18,ma:19}. Ktena \textit{et al.}~\cite{ktena:18} propose to learn a graph similarity metric using a siamese graph convolutional neural network in a supervised fashion, yielding encouraging results in individual subject classification and manifold learning tasks. Similarly, Ma \textit{et al.}~\cite{ma:19} introduce a higher-order siamese graph convolutional neural network for multi-subject brain analysis in health and neuropsychiatric disorders by incorporating higher-order proximity in graph convolutions, with the goal of characterizing the community structure of brain connectivity networks and learning the similarity among magnetic resonance imaging (fMRI) brain connectivity networks extracted from multiple subjects.

While GCNs have been successfully used in the prediction of developmental and brain disorders such as autism spectrum disorder (ASD) and Alzheimer's disease (AD)~\cite{parisot:18,zheng:21,cao:21}, they are prone to the oversmoothing problem, where learned node representations become similar due to repeated graph convolutions as the network depth increases. In other words, when the number of GCN layers increases, the learned node representations tend to converge to indistinguishable feature vectors, resulting in performance degradation and less expressiveness; and hence the model becomes less aware of the graph structure.

In this paper, we aim to address the challenges associated with understanding and diagnosing neurodevelopmental and neurodegenerative brain disorders, such as ASD and AD. We draw inspiration from the biological principles underlying brain function and structure. The human brain, with its intricate network of neurons and their connections, serves as an ideal model for information processing and cognition. By incorporating biologically-inspired aspects into our approach, we aim to capture the fundamental principles that govern brain functioning and leverage them to enhance the understanding and prediction of developmental and brain disorders. To address the challenges associated with increasing network depth and the oversmoothing problem commonly observed in GCNs, we propose an aggregator normalization graph convolutional network (AN-GCN) with skip connections and identity mapping for the detection of neurodevelopmental and neurodegenerative brain disorders by integrating both imaging and non-imaging features into the graph nodes and edges, respectively. We formulate the disease prediction problem as a semi-supervised node classification on population graphs. The main contributions of this work can be summarized as follows:

\begin{itemize}
\item We propose a novel graph convolutional aggregation approach with skip connections and identity mapping for node classification by effectively integrating into the graph both imaging and non-imaging information.
\item We employ an aggregator normalization mechanism for feature propagation in an effort to eliminate bias in minibatch estimation.
\item Our experimental results demonstrate that our model achieves competitive performance compared to robust baseline models on two large benchmark datasets.
\end{itemize}

Unlike existing GCN-based approaches~\cite{parisot:18,zheng:21,cao:21}, our method incorporates both skip connections and identity mapping to maintain the model's awareness of the graph structure and preserve the expressiveness of learned node representations. By adopting a graph sampling-based approach, the proposed aggregator normalization mechanism for feature propagation offers two key benefits. First, it helps eliminate bias in minibatch estimation, ensuring that the model's performance is not skewed by the selection of specific minibatches during training. This leads to more reliable and unbiased parameter estimation. Second, it promotes stability and robustness in feature propagation by reducing the sensitivity to variations in the minibatch data. This enhances the generalization capability of the model, allowing it to perform consistently across different datasets.

The remainder of this paper is organized as follows. In Section 2, we review important relevant work. In Section 3, we present the problem formulation as a semi-supervised node classification task, and then we introduce a two-stage graph convolutional aggregation framework for disease prediction. In the first stage, we construct a population graph comprised of a node set and an edge set with complementary imaging and non-imaging data, respectively. Each graph node represents a subject with an associated feature vector extracted from imaging data, and each edge captures similarities between a pair of subjects with non-imaging data integrated into the edge weight. In the second stage, we design an aggregator normalization graph convolutional network architecture by leveraging skip connections, identity mapping and aggregation in graph sampling. In Section 4, we present experimental results to demonstrate the competitive performance of our approach in comparison with graph-based methods for brain disease prediction. Finally, we conclude in Section 5 and highlight some promising directions for future work.

\section{Related Work}
The basic objective of node classification in populations and diagnosis is to predict the most probable labels of nodes in a population graph, where each subject is represented by a node and each edge encodes the pairwise similarity between a pair of connected nodes. To achieve this objective, various graph convolution based methods have been proposed with the aim of distinguishing between diseased patients and healthy controls by predicting the node labels (i.e., clinical status of subjects). Semi-supervised node classification typically involves a limited number of labeled nodes for model training. The aim is to predict the labels of a vast number of unlabeled nodes by learning a prediction rule from both labeled and unlabeled nodes. This technique improves the model's performance.

\medskip\noindent\textbf{Graph Convolutional Networks.} GCNs have recently become the model of choice in semi-supervised node classification tasks~\cite{kipf:17}. GCN employs a layer-wise propagation rule that utilizes a first-order approximation of spectral graph convolutions. This method updates the feature vector of each node in the graph by computing a weighted sum of the feature vectors of its neighboring nodes. Xu \textit{et al.}~\cite{Bingbing:19} introduce a graph wavelet neural network, which is a GCN-based architecture that uses spectral graph wavelets in lieu of graph Fourier bases to define a graph convolution. Although spectral graph wavelets have the ability to localize graph signals in both spatial and spectral domains, their implementation requires explicit computation of the Laplacian eigenbasis. This results in a high computational complexity, particularly for larger graphs.

While GCNs have shown great promise, achieving state-of-the-art performance in semi-supervised node classification tasks, they are prone to oversmoothing the node features. The neighborhood aggregation scheme utilized by GCN (i.e., graph convolution) is essentially equivalent to performing Laplacian graph smoothing~\cite{Li:18}, which replaces each graph node with the average of its immediate neighbors~\cite{Hamza2007IP,Emad2007GI,Emad2009ISIVP}. Consequently, as the number of network layers increases, applying GCN repeatedly results in increasingly smoother versions of the original node features. This causes the node features in deeper layers to eventually converge to the same value, causing them to become too similar across different classes. Wu \textit{et al.}~\cite{wu:19} introduce a simple graph convolution method by eliminating the nonlinear transition functions between the layers of graph convolutional networks. This results in a linear transformation achieved through powers of the normalized adjacency matrix, augmented with self-loops for all nodes in the graph. However, this simplified approach serves as a low-pass filter that dampens all frequencies except for the zero frequency, leading to oversmoothing. Significant strides have been made toward remedying the problem of oversmoothing in GCNs~\cite{Keyulu:18,Lingxiao:20,chen:20}. Xu \textit{et al.}\cite{Keyulu:18} propose Jumping Knowledge Networks, which incorporate dense skip connections to link each layer of the network with the final layer. This technique helps maintain the locality of node representations and circumvents the problem of oversmoothing. In~\cite{Lingxiao:20}, a normalization layer, which helps avoid oversmoothing by preventing learned representations of distant nodes from becoming indistinguishable, has been proposed. During training, the normalization layer is applied to intermediate layers with the objective of smoothing nodes within the same cluster while preventing smoothing across nodes from different clusters. Chen \textit{et al.}\cite{chen:20} design a deep graph convolutional network that incorporates initial residual and identity mapping. This approach addresses the issue of oversmoothing by augmenting the learnable weight matrix with an identity matrix and utilizing skip connections from the initial feature matrix.

\medskip\noindent{\textbf{Disease Prediction.}} GCNs have recently shown great potential in neuroimaging and computer aided diagnosis, especially in the prediction of brain diseases such as autism spectrum disorder and Alzheimer's disease~\cite{parisot:18,Kazi:19,cosmo:20,cao:21,zheng:21,pan:21,Yao:21,Alzubi2019Indian}. Using a graph convolutional neural network model consisting of a fully convolutional GCN with several hidden layers activated via the Rectified Linear Unit (ReLU) function, Parisot \textit{et al.}~\cite{parisot:18} introduce a disease prediction framework. It involves modeling a population as a graph with nodes representing subjects and edges encoding the similarity between a pair of subjects by combining imaging and non-imaging information in order to improve model classification performance with the goal of distinguishing between patients with autism spectrum disorder and healthy controls, as well as predicting whether a patient with mild cognitive impairment will convert to Alzheimer's disease. The graph nodes are associated with imaging-based features, while non-imaging data is integrated into the edge weights. To learn an adaptive graph representation for GCN learning, Zheng \textit{et al.}~\cite{zheng:21} integrate graph learning and graph convolution to develop an end-to-end multimodal graph learning approach for disease prediction via a multi-modal fusion module, which fuses the features of each modality by leveraging the correlation and complementarity between the modalities. Cao \textit{et al.}~\cite{cao:21} introduce a deep learning model using a multi-layer GCN in conjunction with residual neural networks to tackle the vanishing gradient problem, and the DropEdge technique~\cite{Rong:20} to alleviate overfitting and oversmoothing, which are two major challenges in developing deep GCNs for node classification. Similar to Dropout technique that randomly sets the outgoing edges of hidden units to zero at each update of the training phase, DropEdge can be regarded as an extension of Dropout to graph edges. Inspired by  the Inception network in convolutional neural networks, Kazi \textit{et al.}~\cite{Kazi:19} propose an Inception graph convolutional network for disease prediction tasks with complementary imaging and non-imaging multi-modal data by leveraging spectral convolutions with different kernel sizes, showing improved performance over regular GCN architectures. Cosmo \textit{et al.}~\cite{cosmo:20} present an end-to-end trainable graph learning architecture for dynamic and localized graph pruning with the goal of building a node classification model consisting of few graph convolutional layers, followed by a fully connected layer to predict the patient label. Building upon GCNs, Jiang \textit{et al.}~\cite{Jiang:20} introduce a GCN model with a hierarchical structure designed for learning graph embeddings of the brain network and predicting brain disorders by hierarchically learning deep representations from functional fMRI brain connectivity networks in order to improve classification performance for disease diagnosis. Pan \textit{et al.}~\cite{pan:21} propose a diagnosis classification framework that incorporates self-attention graph pooling and graph convolutional networks by extracting features from the non-Euclidean brain network, as well as fusing both imaging and non-imaging information with the aim of detecting inter-group heterogeneity and intra-group homogeneity regarding brain activities.

While these approaches have shown promising results in brain disease prediction tasks, they are, however, prone to the issue of oversmoothing. Our method differs from existing GCN-based approaches by not only incorporating skip connections and identity mapping to preserve the graph structure and maintain the expressiveness of learned node representations, but also by integrating both imaging and non-imaging features into the graph, mirroring the multidimensional nature of brain-related data. Incorporating skip connections and identity mapping allow our proposed model to retain important features and prevent the collapse of node representations, enabling better discrimination and classification of nodes, particularly in developmental and brain disorder contexts. Inspired by the aggregation mechanism in graph sampling, we also propose an aggregated feature diffusion rule for node features with the aim of leveraging the benefits and insights gained from graph sampling, such as efficient training, enabling the model to make accurate predictions while effectively utilizing computational resources.

\section{Method}
In this section, we introduce our notation and formulate the disease prediction problem as a semi-supervised node classification task on population graphs, which are used to model pairwise relations (edges) between subjects (nodes). Each graph node and edge weight are associated with complementary imaging and non-imaging data, respectively. Then, we present the main building blocks of the proposed network architecture for graph representation learning and semi-supervised node classification.

\subsection{Preliminaries and Problem Statement}
\textbf{Basic Notions.}\quad Consider an undirected graph $\mathcal{G}=(\mathcal{V},\mathcal{E})$, where $\mathcal{V}=\{1,\ldots,N\}$ is the set of $N$ nodes and $\mathcal{E}\subseteq \mathcal{V}\times\mathcal{V}$ is the set of edges. We denote by $\bm{A}=(\bm{A}_{ij})$ an $N\times N$ adjacency matrix (binary or real-valued) whose $(i,j)$-th entry $\bm{A}_{ij}$ is equal to the weight of the edge between neighboring nodes $i$ and $j$, and 0 otherwise. We also denote by $\bm{X}=(\bm{x}_{1},...,\bm{x}_{N})^{\T}$ an $N\times F$ feature matrix of node attributes, where $\bm{x}_{i}$ is an $F$-dimensional row vector for node $i$.

The objective of learning latent node representations in a graph is to learn low-dimensional embeddings that encode both the graph's structural and semantic information. More precisely, the purpose of network/graph embedding is to learn a mapping $\varphi: \mathcal{V}\to\mathbb{R}^{P}$ that maps each node $i$ to a $P$-dimensional vector $\bm{z}_i$, where $P\ll N$. These learned node embeddings can subsequently serve as inputs to learning algorithms for downstream tasks~\cite{Pickup16IJCV,Biasotti16VC}.

\medskip\noindent{\textbf{Problem Statement.}}\quad Semi-supervised learning in a graph involves predicting the labels of nodes that are not labeled, based on the labels of a subset of nodes (or their final output embeddings). More specifically, let $\mathcal{D}_{l}=\{(\bm{z}_i,y_i)\}_{i=1}^{N_l}$ be the set of labeled final output node embeddings $\bm{z}_i\in\mathbb{R}^{P}$ with associated known labels $y_i\in\mathcal{Y}_l$, and $\mathcal{D}_{u}=\{\bm{z}_i\}_{i=N_l+1}^{N_l+N_u}$ be the set of unlabeled final output node embeddings, where $N_l+N_u=N$. The problem of semi-supervised node classification is to learn the parameters $\bg{\theta}$ of a classifier $f_{\bg{\theta}}: \mathcal{V}_{l}\to\mathcal{Y}_l$, where $\mathcal{V}_{l}\subset\mathcal{V}$ is the set of labeled nodes. Then, the aim is to predict the labels of the set $\mathcal{D}_{u}$.

\medskip
It is important to note that for multi-class classification problems, the label of each node $i$ (or its final output embedding $\bm{z}_i$) in the labeled set $\mathcal{D}_{l}$ can be represented as a $C$-dimensional one-hot encoding vector $\bm{y}_{i} \in \{0, 1\}^{C}$, where $C$ is the number of classes with 0 and 1 representing ``healthy'' and ``diseased'' status of the subjects, respectively.

\subsection{Proposed Model}
We now describe our proposed model, a two-stage approach for graph representation learning and semi-supervised node classification. The aim is to learn discriminative node embeddings for computer aided diagnosis. In the first stage, we construct a population graph, which is a vital step in designing a GCN-based prediction model since GCNs rely on the affinity matrix between subjects to update their layer-wise feature propagation rules. Hence, to fully exploit the expressive power of GCNs, an appropriately constructed graph that accurately explains the similarity between subjects is of paramount importance in graph representation learning, especially in computer aided diagnosis tasks. In the second stage, we introduce a disease prediction model by leveraging graph convolutional aggregation in conjunction with skip connections and identity mapping.

\subsubsection{Population Graph Construction}
\begin{figure*}[!ht]
\centering
\includegraphics[scale=.7]{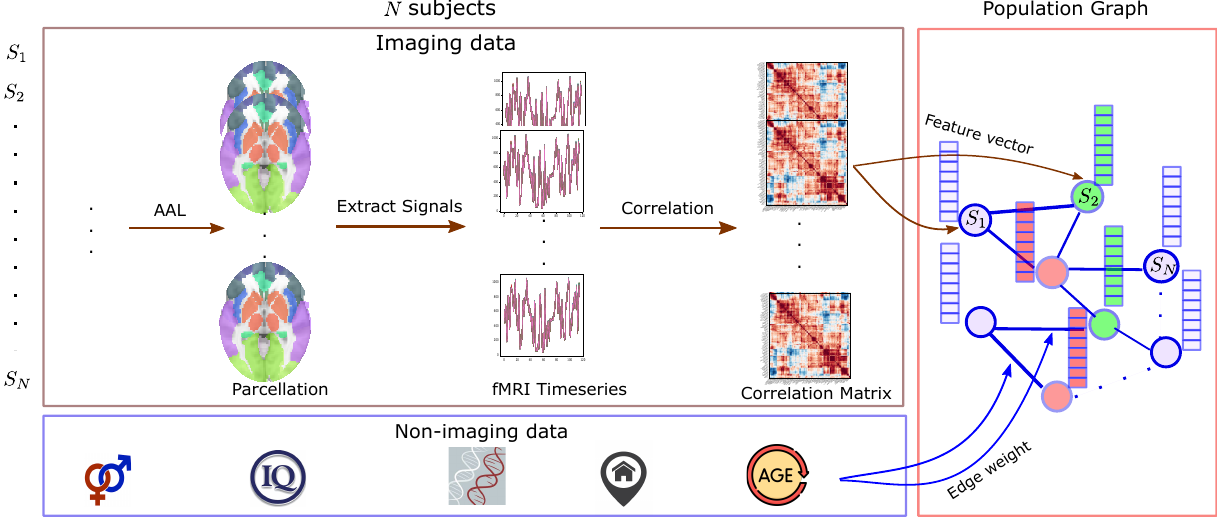}
\caption{Graph construction from $N$ subjects using imaging and non-imaging data. For imaging data, we employ Automated Anatomical Labeling (AAL) to perform brain parcellation.}
\label{Fig:Graph_Construct}
\end{figure*}

Following the population graph construction in graph convolutional networks for disease prediction~\cite{parisot:18}, we also combine both imaging and non-imaging data in our proposed approach. Specifically, we model a population as a graph consisting of nodes representing subjects and edges capturing pairwise similarities between subjects. Each node has a feature vector extracted from imaging data, whereas each edge weight represents phenotypic (i.e., non-imaging) data. The graph construction is shown in Figure~\ref{Fig:Graph_Construct}, where the nodes are associated with imaging-based feature vectors, while phenotypic (non-imaging) information is incorporated as edge weights.

Let $\{M_{1}, \ldots, M_{T}\}$ be a set of $T$ non-imaging phenotypic measures such as a subject's age or gender. The adjacency matrix $\bm{A}=(\bm{A}_{ij})$ of a population graph comprised of $N$ subjects is defined as
\begin{equation}
\bm{A}_{ij} = K(i,j)\sum_{t=1}^{T}d(M_{t}(i),M_{t}(j)),
\label{Eq:adjacency_matrix}
\end{equation}
where $K(i,j)=\text{similarity}(S_i,S_j)$ denotes a kernel similarity between subjects $S_i$ and $S_j$ (i.e., edge weight between graph nodes $i$ and $j$), and $d$ is a pairwise distance between phenotypic measures. The kernel similarity measure $K(i,j)$ is given by
\begin{equation}
K(i,j)= \exp \Big(- \frac{\rho(\bm{x}_{i},\bm{x}_{j})^{2}}{2\sigma^{2}} \Big),
\end{equation}
where $\sigma$ is a smoothing parameter, which determines the width of the kernel, and $\rho$ is the correlation distance between feature vectors $\bm{x}_{i}$ and $\bm{x}_{j}$ for nodes $i$ and $j$, respectively,
\begin{equation}
\rho(\bm{x}_{i},\bm{x}_{j}) = 1-\frac{(\bm{x}_{i}-\bar{\bm{x}}_{i})(\bm{x}_{j}-\bar{\bm{x}}_{j})^{\T}}{\Vert \bm{x}_{i}-\bar{\bm{x}}_{i}\Vert \Vert \bm{x}_{j}-\bar{\bm{x}}_{j}\Vert},
\end{equation}
with $\bar{\bm{x}}_{i}=(\bm{x}_{i}\bm{1}/N)\bm{1}^{\T}$ and $\bar{\bm{x}}_{j}=(\bm{x}_{j}\bm{1}/N)\bm{1}^{\T}$ denoting row vectors whose elements are all equal to the mean of the components of $\bm{x}_{i}$ and $\bm{x}_{j}$, respectively, and $\bm{1}$ is an $N$-dimensional column vector of all ones.

The pairwise distance between phenotypic measures is defined depending on the kind of phenotypic data incorporated in the graph. Most phenotypic data can be classified into two main categories: qualitative (e.g., subject's gender) and quantitative (e.g., subject's age). For qualitative data, the distance measure is defined as
\begin{equation}
d(M_{t}(i),M_{t}(j))=
\left\{\begin{array}{lll}
1 & \mbox{if } M_{t}(i) = M_{t}(j) \cr
0 & \text{otherwise}.
\end{array}
\right.
\end{equation}
and for quantitative data, it is given by
\begin{equation}
d(M_{t}(i),M_{t}(j))=
\left\{\begin{array}{lll}
1 & \mbox{if } \lvert M_{t}(i) - M_{t}(j)\rvert < \tau, \cr
0 & \text{otherwise}.
\end{array}
\right.
\end{equation}
where $\tau$ is a given threshold.

\subsubsection{Disease Prediction Model}
Graph convolutional networks learn a new feature representation for each node such that nodes with the same labels have similar features~\cite{kipf:17}.

\medskip\noindent{\textbf{Feature Diffusion.}}\quad We denote by $\tilde{\bm{A}}=\bm{A}+\bm{I}$ the adjacency matrix with self-added loops, where $\bm{I}$ is the identity matrix. The layer-wise feature diffusion rule of an $L$-layer GCN is given by
\begin{equation}
\bm{S}^{(\ell)}=\hat{\bm{A}}\bm{H}^{(\ell)}, \quad \ell=0,\dots,L-1,
\end{equation}
where $\hat{\bm{A}}=\tilde{\bm{D}}^{-\frac{1}{2}}\tilde{\bm{A}}\tilde{\bm{D}}^{-\frac{1}{2}}$ is the normalized adjacency matrix with self-added loops, $\tilde{\bm{D}}=\op{diag}(\tilde{\bm{A}}\bm{1})$ is the diagonal degree matrix, and $\bm{H}^{(\ell)}\in\mathbb{R}^{N\times F_{\ell}}$ is the input feature matrix of the $\ell$-th layer with $F_{\ell}$ feature maps. The input of the first layer is the original feature matrix $\bm{H}^{(0)}=\bm{X}$.

\medskip\noindent{\textbf{Aggregated Feature Diffusion.}}\quad Inspired by the aggregation mechanism in graph sampling~\cite{zeng:19}, we define a layer-wise aggregated feature diffusion rule for node features in the $\ell$-th layer as follows:
\begin{equation}
\bm{S}^{(\ell)}=(\hat{\bm{A}}\odot\bg{\Gamma})\bm{H}^{(\ell)},
\label{Eq:AFD}
\end{equation}
where $\odot$ denote element-wise matrix multiplication, and $\bg{\Gamma}=(\gamma_{ij})$ is an $N\times N$ aggregation matrix. Each element $\gamma_{ij}$ is an aggregator normalization constant given by
\begin{equation}
\gamma_{ij}=\frac{C_{i}}{C_{ij}},
\end{equation}
where $C_{i}$ and $C_{ij}$ denote the number of times the node $i\in\mathcal{V}$ or edge $(i,j)\in\mathcal{E}$ appears in the subgraphs of $\mathcal{G}=(\mathcal{V},\mathcal{E})$, respectively. These subgraphs are obtained by running the GraphSaint sampler~\cite{zeng:19} repeatedly before the training starts. The minibatches constructed through graph sampling contain a fixed number of well-connected nodes in all layers, ensuring consistent and effective learning.

\medskip\noindent{\textbf{Learning Node Embeddings.}}\quad Motivated by the good performance of graph sampling and identity mapping in alleviating the oversmoothing problem in graph representation learning~\cite{zeng:19,wu:19,klicpera:19,chen:20,chu:21,paetzold:21}, we propose an aggregator normalization graph convolutional network (AN-GCN) by leveraging aggregation in graph sampling, as well as skip connections and identity mapping. The output feature matrix $\bm{H}^{(\ell+1)}$ of our proposed AN-GCN model is obtained by applying the following layer-wise propagation rule
\begin{equation}
\begin{split}
\bm{H}^{(\ell + 1)} &= \sigma\bigg(\big(1 - \alpha_{\ell}\big)(\hat{\bm{A}}\odot\bg{\Gamma})\bm{H}^{(\ell)} \\ &\hspace*{.9cm}+\beta_{\ell}(\hat{\bm{A}}\odot\bg{\Gamma})\bm{H}^{(\ell)} \big(\bm{I} + \bm{W}^{(\ell)}\big)\\
&\hspace*{.9cm}+\alpha_{\ell}\bm{X} + \beta_{\ell}\bm{X}\big(\bm{I}+ \bm{W}^{(\ell)}\big)\bigg),
\end{split}
\label{Eq:K_layers_aggregation_model}
\end{equation}
where $\alpha_{\ell}$ and $\beta_{\ell}$ are nonnegative hyper-parameters in the interval $[0,1]$ and are often fine-tuned via grid search, $\bm{W}^{(\ell)}$ is a trainable weight matrix at the $\ell$-layer, and $\sigma(.)$ is a point-wise non-linear activation function such as $\textnormal{ReLU}(.) = \textnormal{max}(0,.)$. By incorporating skip connections, the resulting representation of each node includes a minimum proportion of feature information from the input layer, as determined by $\alpha_{\ell}$. Additionally, the use of identity mapping serves a dual purpose of imposing regularization on the weight matrix to prevent over-fitting, as well as being advantageous for semi-supervised learning scenarios in which the available training data is limited~\cite{li:17}.

\medskip\noindent{\textbf{Model Prediction.}} The embedding matrix $\bm{Z}=\bm{H}^{(L)}$ of the last layer of AN-GCN contains the final output node embeddings, and captures the neighborhood structural information of the graph within $L$ hops. This final node representation can be used as input for node classification. To this end, we apply a softmax classifier as follows:
\begin{equation}
\hat{\bm{Y}}=\text{softmax}(\bm{Z}),
\end{equation}
where $\hat{\bm{Y}}\in\mathbb{R}^{N\times C}$ is the matrix of predicted labels for graph nodes, and $C$ is the total number of classes. The softmax classifier is a generalization of the binary logistic regression classifier to multiple classes, and as the name suggests it uses the softmax function that turns a vector of $C$ real-valued class scores into a vector of $C$ normalized positive scores that sum to 1. In other words, the softmax classifier returns probability scores for all classes.

\medskip\noindent{\textbf{Model Training.}} For semi-supervised multi-class classification, the neural network weight parameters are learned by minimizing the cross-entropy loss function
\begin{equation}
\mathcal{L}=-\sum_{i\in \mathcal{V}_{l}}\sum_{c=1}^{C} \bm{Y}_{ic} \log\hat{\bm{Y}}_{ic},
\end{equation}
over the set $\mathcal{V}_{l}$ of all labeled nodes, where $\bm{Y}_{ic}$ is equal 1 if node $i$ belongs to class $c$, and 0 otherwise; and $\hat{\bm{Y}}_{ic}$ is the $(i,c)$-element of the matrix $\hat{\bm{Y}}$ from the softmax function, i.e., the probability that the network associates the $i$-th node with class $c$. During training, the network parameters are updated using the Adam optimizer~\cite{kingma2014adam}.

\begin{figure*}[!htb]
\centering
\includegraphics[scale=.8]{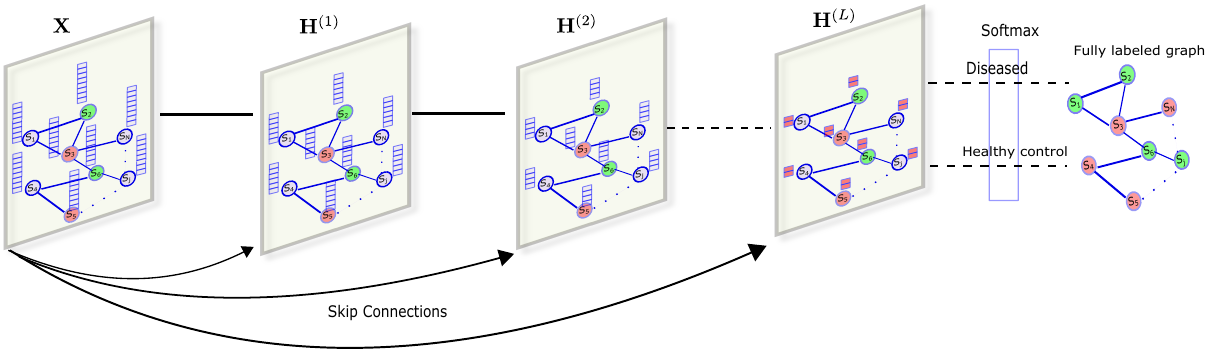}
\caption{Schematic layout of the proposed AN-GCN architecture.}
\label{Fig:Pipeline_model}
\end{figure*}

\section{Experiments}
In this section, we conduct several experiments to assess the performance of the proposed AN-GCN model on two standard datasets for disease prediction. More specifically, we address the disease prediction problem as a semi-supervised node classification task, and the goal is predict the label (i.e., clinical status) of a test node (i.e., subject) in a population graph as diseased or healthy, where only a small number of nodes are labeled. The effectiveness of our model is validated through experimental comparison with strong baseline methods. While presenting and analyzing our experimental results, we aim to answer the following main research questions (RQs):

\begin{itemize}
\item \textbf{RQ1:} How does AN-GCN perform in comparison with state-of-the-art disease prediction models?
\item \textbf{RQ2:} How does AN-GCN alleviate the oversmoothing problem?
\item \textbf{RQ3:} What is the effect of hyperparameters on the performance of AN-GCN?
\end{itemize}

\subsection{Experimental Setup}
The experimental setup of our study encompasses several key components that are essential for evaluating the performance and effectiveness of our proposed approach. In this section, we provide an overview of the datasets used, the data processing steps undertaken, the performance evaluation metrics employed, the baseline methods for comparison, and the implementation details. By detailing these aspects, we aim to ensure transparency, reproducibility, and comprehension of the experimental methodology and results interpretation.

\subsubsection{Datasets}
We evaluate the proposed model on two large datasets, namely ABIDE and ADNI.
\begin{itemize}
\item \textbf{ABIDE Dataset:}\quad The Autism Brain Imaging Data Exchange (ABIDE)\footnote{http://preprocessed-connectomes-project.org/abide/}\cite{DiMartino14Abide} initiative aggregates resting-state functional magnetic resonance imaging (rs-fMRI) and phenotypic data of 1112 subjects from various international brain imaging laboratories. We select a set of 871 subjects, consisting of 403 ASD patients and 468 healthy controls (HCs). As a result of the different acquisition sites, the ABIDE dataset is heterogeneous, and the aim is to separate ASD subjects from healthy controls.

\item \textbf{ADNI Dataset:}\quad The Alzheimer's Disease Neuroimaging Initiative (ADNI)\footnote{http://adni.loni.usc.edu/}\cite{Peterson10ADNI} is a North American multisite study designed to develop clinical, neuroimaging techniques, biochemical and genetic biomarkers for the early detection and tracking of patients with Alzheimer's disease (AD), as well as subjects with mild AD, normal subjects, and subjects with mild cognitive impairment (MCI). ADNI has recruited more than 1700 adults, aged 55 to 90 years, from over 50 sites across the United States and Canada for its four studies (ADNI-1, 2, 3 and -GO). We select a set of 573 participants, comprised of 402 HC individuals and 171 MCI subjects. The aim is to predict whether an MCI subject will convert to AD.
\end{itemize}

\subsubsection{Data Preprocessing}
For fair comparison, we follow the same data preprocessing procedure laid out in the GCN baseline~\cite{parisot:18}. For preprocessing of the ABIDE dataset, we use the Configurable Pipeline for the Analysis of Connectomes (C-PAC)~\cite{Craddock13CPAC}, which includes skull stripping, slice timing correction, motion correction, global mean intensity normalization, nuisance signal regression, band-pass filtering (0.01-0.1Hz), and registration of fMRI images to a standard anatomical space. Then, the mean timeseries for a set of cortical and subcortical regions of interest (ROIs) extracted from the Harvard Oxford atlas are computed and standardized using z-score normalization to ensure the timeseries distributions have mean zero mean and unit variance. The goal of z-score normalization is to transform timeseries to be on a similar scale in an effort to improve the performance and training stability of the model. Subsequently, we compute $N$ connectivity matrices using the Pearson's correlation coefficient between the representative rs-fMRI timeseries of each ROI in the Harvard Oxford atlas. Since z-scores are not necessarily normally distributed, we apply Fisher z-transformation, which is the inverse hyperbolic tangent function that converts Pearson's correlation coefficient to a normally distributed variable. In other words, the correlation matrices are Fisher transformed in order to convert the skewed distribution of the correlation coefficient into a distribution that is approximately normal. It is also worth pointing out that the variance of the Fisher transformed distribution is independent of the correlation, whereas the variance of the sampling distribution of the correlation coefficient depends on the correlation. For the edge weights of the population graph on the ABIDE datset, we incorporate the subject's gender, age and acquisition site as as phenotypic measures. For the ADNI dataset, we parcellate each 3D brain volume into $N$ ROIs using Automated Anatomical Labeling (AAL)~\cite{Tzourio02AAL}, followed by computing the connectivity matrices between timeseries. The edge weights of the population graph on the ADNI dataset consist of the subject's gender and age as phenotypic measures. Since the correlation matrix is symmetric, it suffices to use either its upper or lower triangular part. Hence, we take the upper triangular part and vectorize it to obtain a feature vector whose dimension is then further reduced using recursive feature elimination via a ridge classifier.

\subsubsection{Performance Evaluation Metrics}
To evaluate the performance of a classification model, it is typically applied to folds within a 10-fold cross-validation setup, where each fold represents a subset of the data with known target values. The model's predictions are then compared to the actual known values. We use Accuracy (Acc), Area Under Curve (AUC), Recall, Precision, F1 score, Matthews Correlation Coefficient (MCC), and Cohen's kappa ($\kappa$) as evaluation metrics, which are defined as
$$
\text{Acc} = \frac{\text{TP + TN}}{\text{TP + TN + FP + FN}},
$$
$$
\text{Recall} = \frac{\text{TP}}{\text{TP + FN}},
$$
$$
\text{Precision} = \frac{\text{TP}}{\text{TP + FP}},
$$
$$
\text{F1} = \frac{2 \times \text{Precision} \times \text{Recall}}{\text{Precision}+\text{Recall}},
$$
$$
\text{MCC} = \frac{\text{TP} \times \text{TN} - \text{FP} \times \text{FN}}{\sqrt{(\text{TP + FP})(\text{TN + FP})(\text{TP + FN})(\text{TN + FN})}},
$$
and
$$
\kappa = \frac{2\times(\text{TP}\times\text{TN}-\text{FP}\times\text{FN})}{(\text{TP}+\text{FP})\times(\text{TN}+\text{FP}) + (\text{TP}+\text{FN})\times(\text{TN}+\text{FN})},
$$
where TP, FP, TN and FN denote number of true positives, false positives, true negatives and false negatives, respectively.

The F1-score is defined as the harmonic mean of precision and recall. The harmonic mean is more intuitive than the arithmetic mean when computing a mean of ratios. The F1-score will only be high if both precision and recall have high values. This is due to the fact that the harmonic mean of two numbers is always closer to their minimum.

We also use AUC, the area under the receiving operating characteristic (ROC) curve, as a metric. AUC summarizes the information contained in the ROC curve, which plots the true positive rate versus the false positive rate, at various thresholds. Larger AUC values indicate better performance at distinguishing between healthy and diseased subjects. An uninformative classifier has an AUC equal to 50\% or less. An AUC of 50\% corresponds to a random classifier (i.e., for every correct prediction, the next prediction will be incorrect), whereas an AUC score smaller than 50\% indicates that the classifier performs worse than a random one.

\subsubsection{Baseline Methods}
We evaluate the performance of the proposed AN-DGCN model against various graph convolutional based methods for computer aided diagnosis, including GCN for disease prediction~\cite{parisot:18}, multi-modal graph learning (MMGL) for disease prediction~\cite{zheng:21}, DeepGCN for autism spectrum
disorder identification from multi-site resting-state data~\cite{cao:21}, InceptionGCN for disease prediction~\cite{Kazi:19}, latent-graph learning (LGL) for disease prediction~\cite{cosmo:20}, edge-variational graph convolutional network (EV-GCN) for uncertainty-aware disease prediction~\cite{Huang:20}, down-sampling and multi-modal learning (DS-MML) for identifying autism spectrum disorder~\cite{pan:21}, hierarchical graph convolution network (HI-GCN) for brain disorders prediction~\cite{Jiang:20}, brain connectivity via graph convolution network (BN-GCN) for Alzheimer's Disease~\cite{gu:21}, and mutual multi-scale triplet graph convolutional network (MMTGCN) for brain disorders classification~\cite{Yao:21}. We also compare our model to logistic regression, gradient boosting, and tensor-train, high-order pooling and semi-supervised learning-based generative adversarial network (THS-GAN)~\cite{yu:21}.

\subsubsection{Implementation Details}
\begin{figure}[!htb]
\centering
\includegraphics[scale=.525]{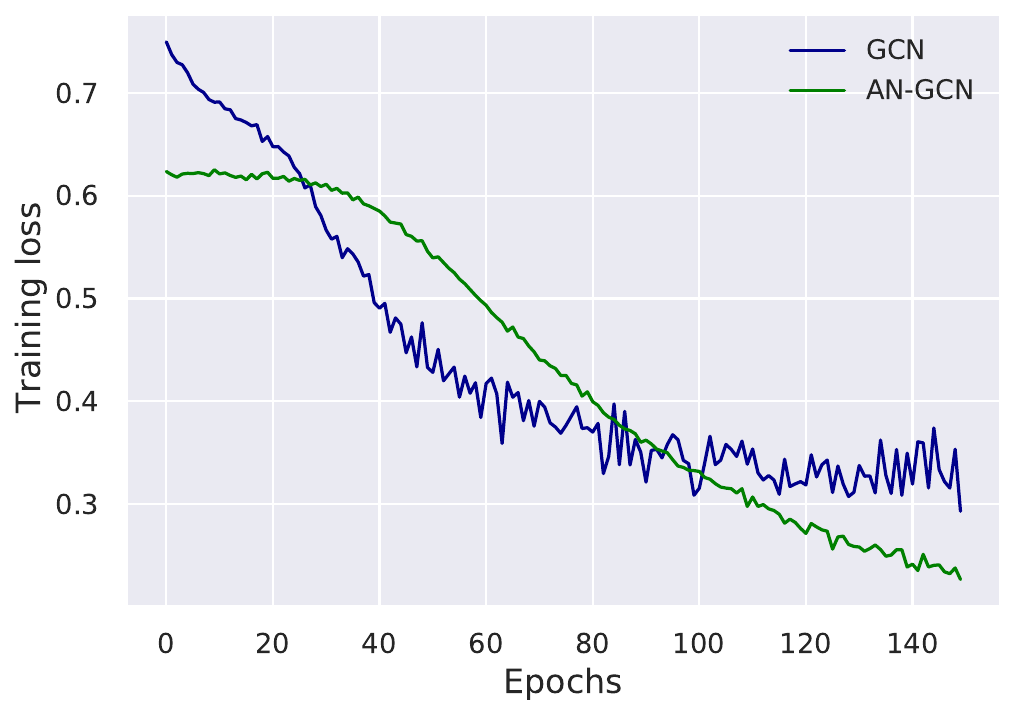}
\caption{Model training history comparison between GCN and our AN-GCN model on the ABIDE dataset.}
\label{Fig:TrainingErros_ABIDE}
\end{figure}

All experiments are carried out on a Linux workstation running Intel(R) CPU 2.40 GHz and 128-GB RAM with an V100-SXM2 16-GB GPU. The proposed model is implemented in PyTorch and trained for 150 and 100 epochs on the ABIDE and ADNI datasets, respectively, using Adam optimizer with a learning rate of $10^{-3}$. The values of the hyperparameters $\alpha_{\ell}$ and $\beta_{\ell}$ are set to 0.1 and 0.3 for the ABIDE dataset, and 0.1 and 0.2 for the ADNI dataset, respectively, via grid search with cross-validation on the training set. We use a stratified 10-fold cross-validation strategy. We also set the number of layers for our model to $L=10$. The training is stopped when the validation loss does not decrease after 10 consecutive epochs. The values of the cross-entropy metric are recorded at the end of each epoch on the training set. The performance comparison plots of AN-GCN and GCN over training epochs on the training set of the ABIDE dataset are visualized in Figure~\ref{Fig:TrainingErros_ABIDE}, which shows that both models yield comparable training loss values. However, as the number of epochs increases, our model yields lower training loss values, indicating better predictive accuracy.

\subsection{Experimental Results and Analysis}
In order to answer \textbf{RQ1}, we report the classification performance of AN-GCN and baseline methods in Table~\ref{Table:Performance_Comparison_ABIDE} using seven evaluation metrics, including average accuracy, AUC and F1-score. Each metric is averaged across all test samples. As can be seen, the results show that our model outperforms all the baseline methods on the ABIDE dataset, achieving relative improvements of 50.33\%, 34.10\%, 50.33\%, 12.25\% and 50.33\% over GCN in terms of accuracy, AUC, F1-score, recall and precision, respectively. The relative improvements over GCN are significant in terms of $\kappa$ and MCC. Compared to the strongest baseline, our model outperforms DS-MML by relative improvements of 10.60\%, 5.28\% and 11.70\% in terms of accuracy, AUC and recall, respectively.

\begin{table*}[!htb]
\centering
\caption{Performance comparison of our model and baseline methods on the ABIDE dataset using various evaluation metrics (\%). Boldface numbers indicate the best classification performance.}
\medskip
\label{Table:Performance_Comparison_ABIDE} 
\begin{tabular}{llcccccccc}
\toprule
                        & Accuracy     & AUC          & F1-score   & Recall	    & Precision  & $\kappa$	     & MCC	\\
Method                  &              &              &            &            & 	         &	             & \\
\midrule
Logistic Regression     & 61.03        & 68.05	      & 70.19      & 88.42      & 58.4       & 19.85         & 25.13 \\
Gradient Boosting       & 59.97	       & 62.04	      & 62.24      & 63.48	    & 61.34		 & 19.53	     & 19.67  \\
GCN~\cite{parisot:18}   & 64.63        & 72.23        & 64.63      & 86.33      & 64.63      & 26.64         & 30.16  \\
InceptionGCN~\cite{Kazi:19} & 72.69    & 72.81        & 79.27      &-           &-           &-              &-      \\
EV-GCN~\cite{Huang:20}  & 80.83        & 84.98        & 81.24      &-           &-           &-              &-      \\
LGL~\cite{cosmo:20}     & 84.69        & 84.46        &-           &-           &-           &-              &-      \\
HI-GCN~\cite{Jiang:20}  & 66.50        & 72.10        &-           & 65.30      &-           &-              &-      \\
MMGL~\cite{zheng:21}    & 86.95        & 86.84        &-           &-           &-           &-              &-      \\
DeepGCN~\cite{cao:21}   & 73.71        & 75.20        & 69.68      &-           &-           &-              &-      \\
DS-MML~\cite{pan:21}    & 87.62        & 92.00        &-           & 86.76      &            &-              &-      \\
\midrule
AN-GCN (Ours)         & \textbf{96.91}   & \textbf{96.86}   & \textbf{97.16} & \textbf{96.91} & \textbf{97.00} &\textbf{93.76} & \textbf{93.86}\\
\bottomrule
\end{tabular}
\end{table*}

Similarly, we report the performance comparison results of our model and baseline methods on the ADNI dataset in Table~\ref{Table:Performance_Comparison_ADNI}, which also shows that AN-GCN performs better than all the competing baselines. Our model yields relative improvements of 15.61\%, 8.66\%, 14.13\% and 15.70\% over GCN in terms of accuracy, AUC, F1-score and precision, respectively. Moreover, AN-GCN significantly outperforms GCN in terms of recall, $\kappa$ and MCC. In addition, our model outperforms the strongest baseline (i.e., BCN-GCN) by relative improvements of 5.76\% and 4.36\% in terms of accuracy and AUC, respectively.

\begin{table*}[!htb]
\centering
\caption{Performance comparison of our model and baseline methods on the ADNI dataset using various evaluation metrics (\%). Boldface numbers indicate the best classification performance.}
\medskip
\label{Table:Performance_Comparison_ADNI}
\begin{tabular}{llcccccccc}
\toprule
                            & Accuracy  & AUC        & F1-score     & Recall	& Precision & $\kappa$	     & MCC	\\
Method                      &           &            &              &           & 	        &	             & \\
\midrule
Logistic Regression         & 58.71		& 51.61	     & 68.80        & 58.71	    & 59.67	    & 04.48	         & 04.04  \\
Gradient Boosting           & 65.21		& 68.57	     & 65.21        & 71.83	    & 65.21	    & 29.52	         & 29.95 \\
GCN~\cite{parisot:18}       & 84.98     & 89.32      & 84.89        & 58.82     & 84.98     & 60.37          & 62.58  \\
HI-GCN~\cite{Jiang:20}      & 75.40     & 75.60      &-             & 66.40     &-           &-              &-      \\
BCN-GCN~\cite{gu:21}        & 92.90     & 93.00      & -            &-          &-          &-               &-      \\
MMTGCN~\cite{Yao:21}        & 86.00     & 90.30      & -            & 86.90     &-          &-               &-      \\
THS-GAN~\cite{yu:21}        & 85.71     & 85.35      & 87.27        & 88.89     & 85.71     &-               &-      \\
\midrule
AN-GCN (Ours)               &\textbf{98.25} & \textbf{97.06} &\textbf{96.89}    &\textbf{98.25} &\textbf{98.33} &\textbf{95.68}  &\textbf{95.84}  \\
\bottomrule
\end{tabular}
\end{table*}

\begin{figure}[!htb]
\centering
\includegraphics[scale=.525]{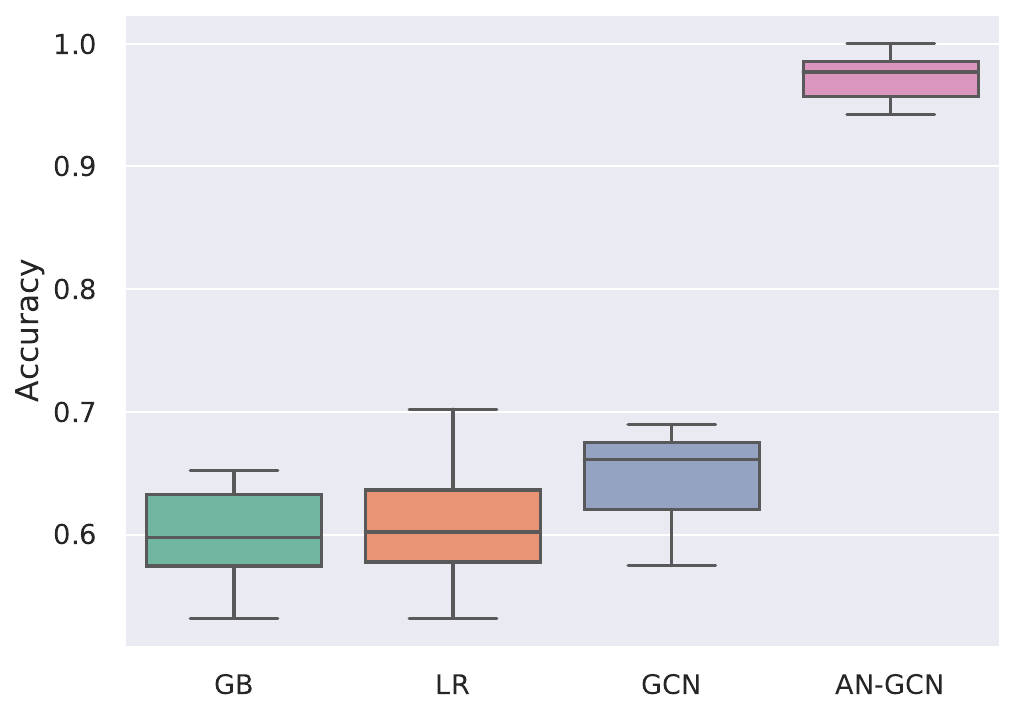}\\
\includegraphics[scale=.525]{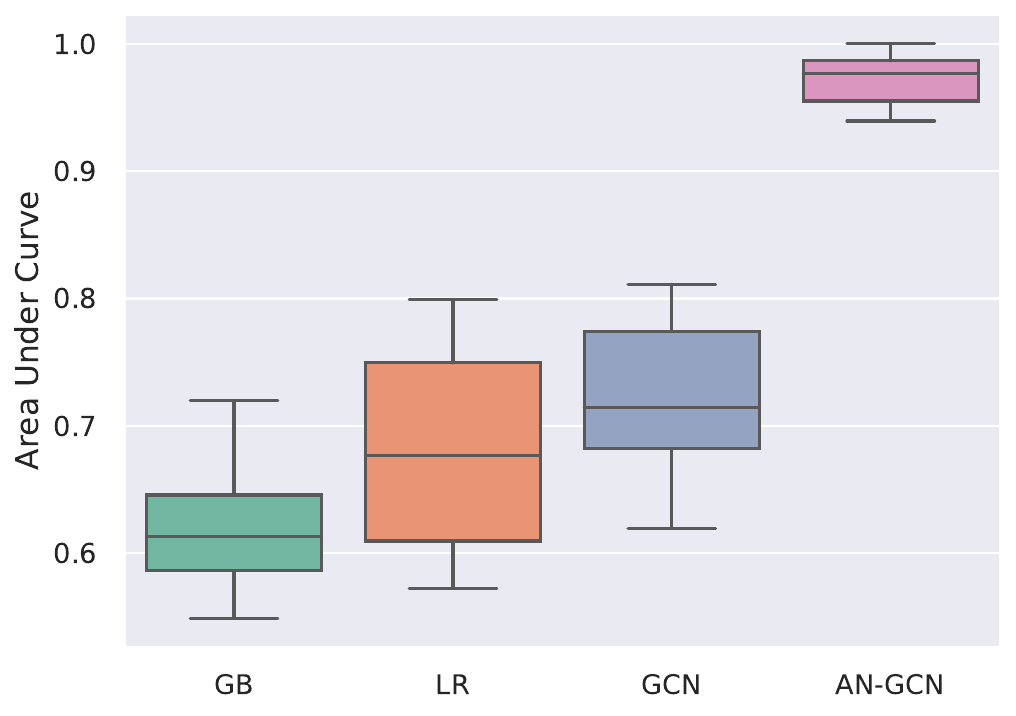}
\caption{Comparative box plots of our model and baseline methods on the ABIDE dataset using accuracy and AUC scores over all cross-validation folds.}
\label{Fig:ABIDE_ACC}
\end{figure}

\begin{figure}[!htb]
\centering
\includegraphics[scale=.525]{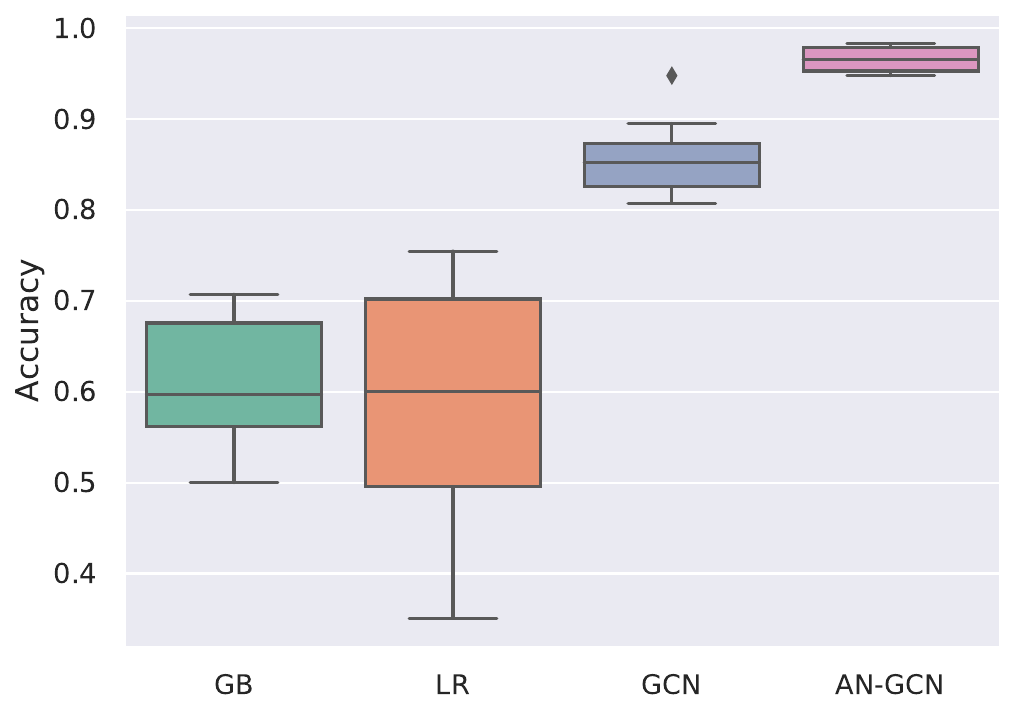}\\
\includegraphics[scale=.525]{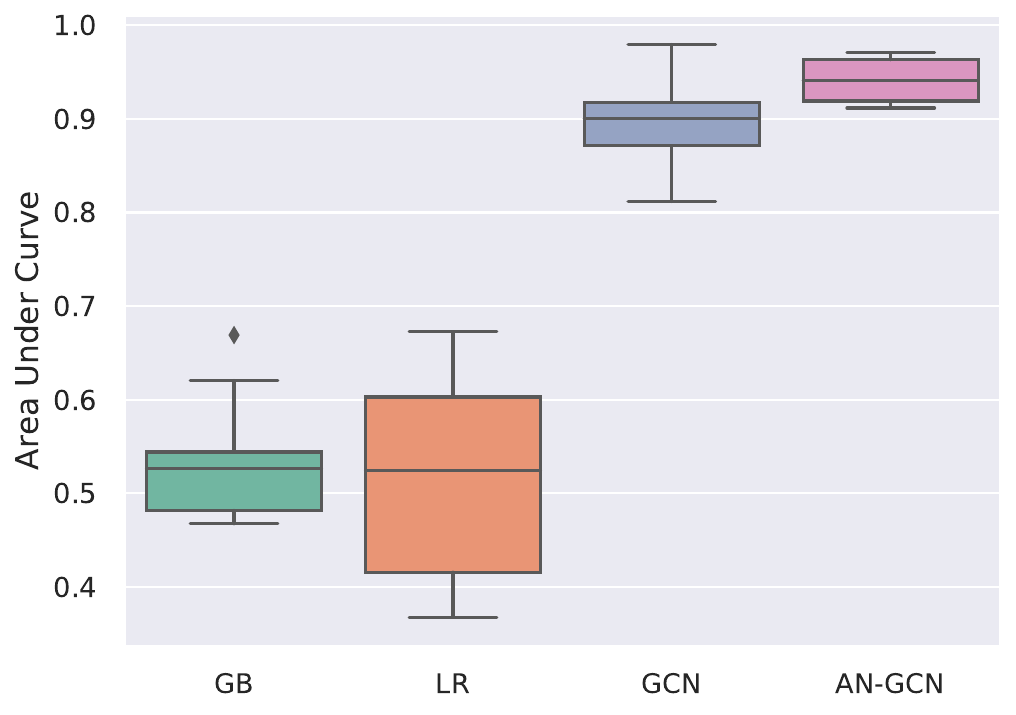}
\caption{Comparative box plots of our model and baseline methods on the ADNI dataset using accuracy and AUC scores over all cross-validation folds.}
\label{Fig:ADNI_ACC}
\end{figure}

In order to visually compare the performance of the proposed model to the baseline methods, we use box plots across all the folds on the ABIDE and ADNI dataset using accuracy and AUC as evaluation metrics, as shown in Figures~\ref{Fig:ABIDE_ACC} and~\ref{Fig:ADNI_ACC}. A box plot is a simple method for graphically depicting groups of numerical data through their quartiles, and it is commonly used to assess and compare the shape, central tendency, and variability of sample distributions, as well as to identify outliers. The box and whiskers show how the data is spread out. On each box, the central line represents the median, and the bottom and top edges of the box indicate the first and third quartile, respectively. The whiskers extend from the edges of the box to the lower and upper inner fences to show the range of the data. The fences are defined in terms of the inter-quartile range, and any value that falls outside the fences in considered as an outlier.

Figure~\ref{Fig:ABIDE_ACC} shows that our model outperforms the competing baselines in terms of the accuracy and AUC metrics for all the 10-folds. The higher the accuracy and AUC scores, the better the model distinguishes between patients suffering from ASD and healthy controls. As can be seen in Figure~\ref{Fig:ABIDE_ACC}, the distribution of our model has less variability than GCN, gradient boosting and logistic regression in autism spectrum disorder prediction tasks. For instance, the median accuracy score for AN-GCN on the ABIDE dataset indicates significant difference in performance between our model and the three baseline methods. In addition, the box for AN-GCN is short, meaning that the accuracy values consistently hover around the average accuracy. However, the boxes for three baselines are taller, implying variable accuracy and AUC values compared to AN-GCN. We can also observe that the whisker is longer on the lower end of the box for GCN, indicating the distribution of both accuracy and AUC values is negatively skewed. For our AN-GCN model, the whisker lengths are short and roughly of the same length, indicating lower standard deviation and data symmetry, respectively.

Similarly, the box plots shown in Figures~\ref{Fig:ADNI_ACC} indicate that our AN-GCN model outperforms the three baseline methods on the ADNI dataset in terms of both accuracy and AUC metrics. Interestingly, the box plot for GCN exhibits an outlier for accuracy values, as well as longer whiskers for AUC values. In addition, the box for the logistic regression is much taller than the other methods, indicating high variability in accuracy and AUC values. Gradient boosting also exhibits an outlier for AUC values.

We also evaluate the performance of our model against competing baselines using the precision-recall (PR) and receiver operating characteristic (ROC) curves on both ABIDE and ADNI datasets. The PR curve summarizes the trade-off between the true positive rate and the positive predictive value for a predictive model using different probability thresholds. Precision is a measure of result relevancy, while recall is a measure of how many truly relevant results are returned. A high area under the PR curve represents both high recall and high precision, where high precision relates to a low false positive rate, and high recall relates to a low false negative rate. Moreover, a PR curve that is closer to the upper left indicates a better performance. On the other hand, the ROC curve summarizes the trade-off between the true positive rate and false positive rate for a predictive model using different probability thresholds. The area under the ROC curve (AUC) is a measure of discrimination in the sense that a model with a high AUC suggests that the model is able to accurately predict the value of an observation's response. Moreover, an ROC curve that is closer to the upper right indicates a better performance (i.e., true positive rate is higher than false positive rate).

\begin{figure}[!htb]
\centering
\includegraphics[scale=.525]{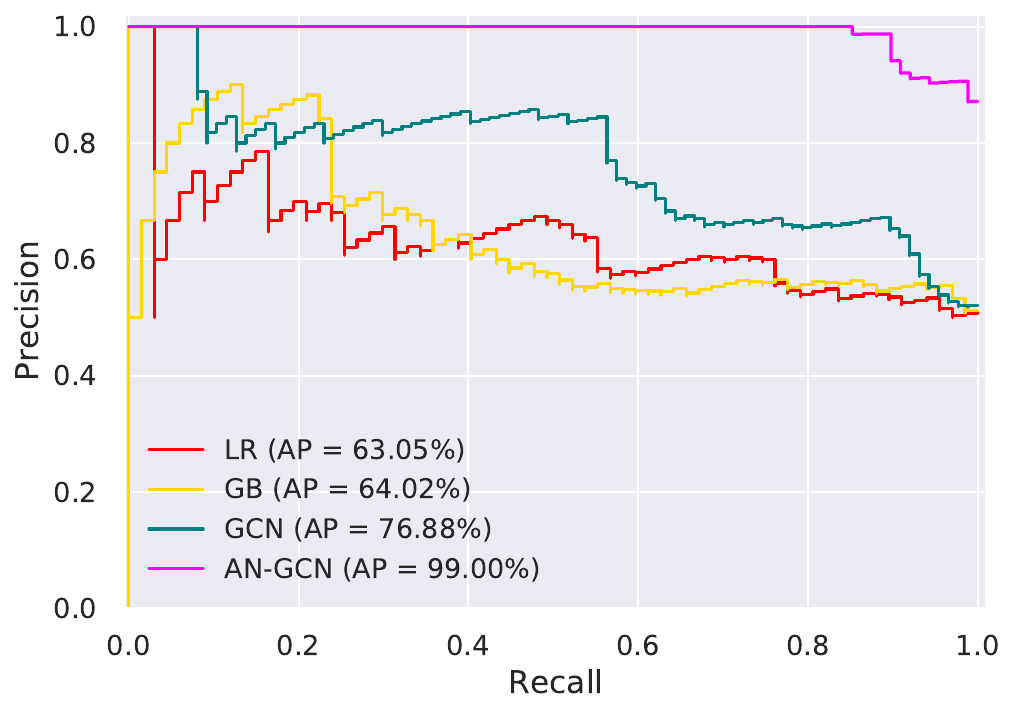}\\
\includegraphics[scale=.525]{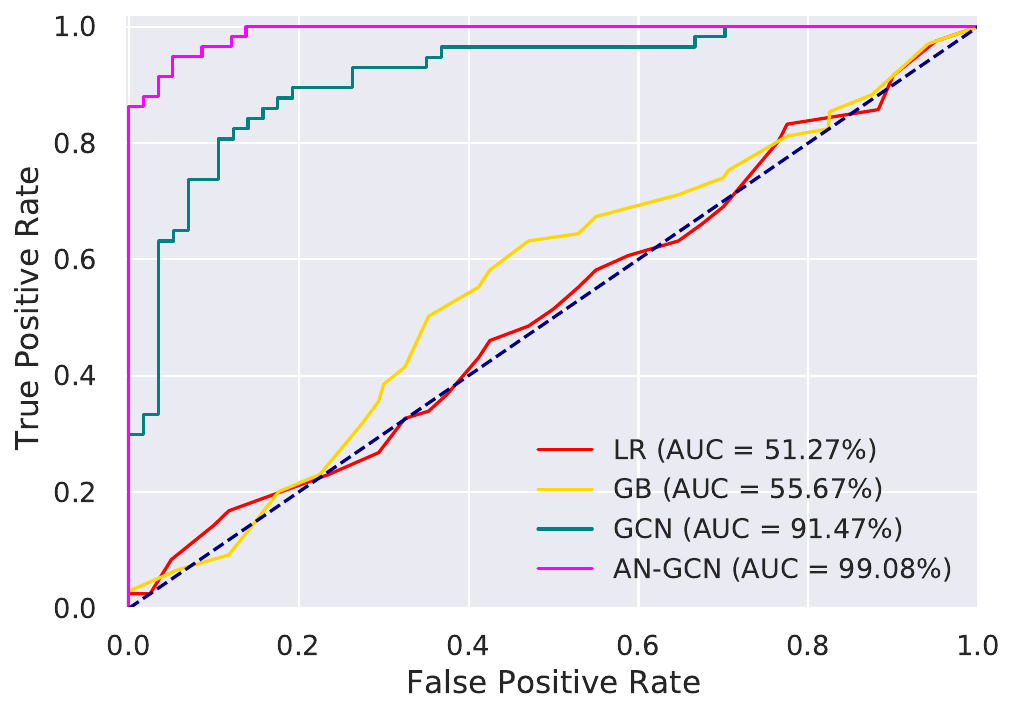}
\caption{Precision-Recall and ROC curves of our model and baseline methods on the ABIDE dataset. Average precision (AP) and AUC values are enclosed in parentheses.}
\label{Fig:ABIDE_recall_precision}
\end{figure}

\begin{figure}[!htb]
\centering
\includegraphics[scale=.525]{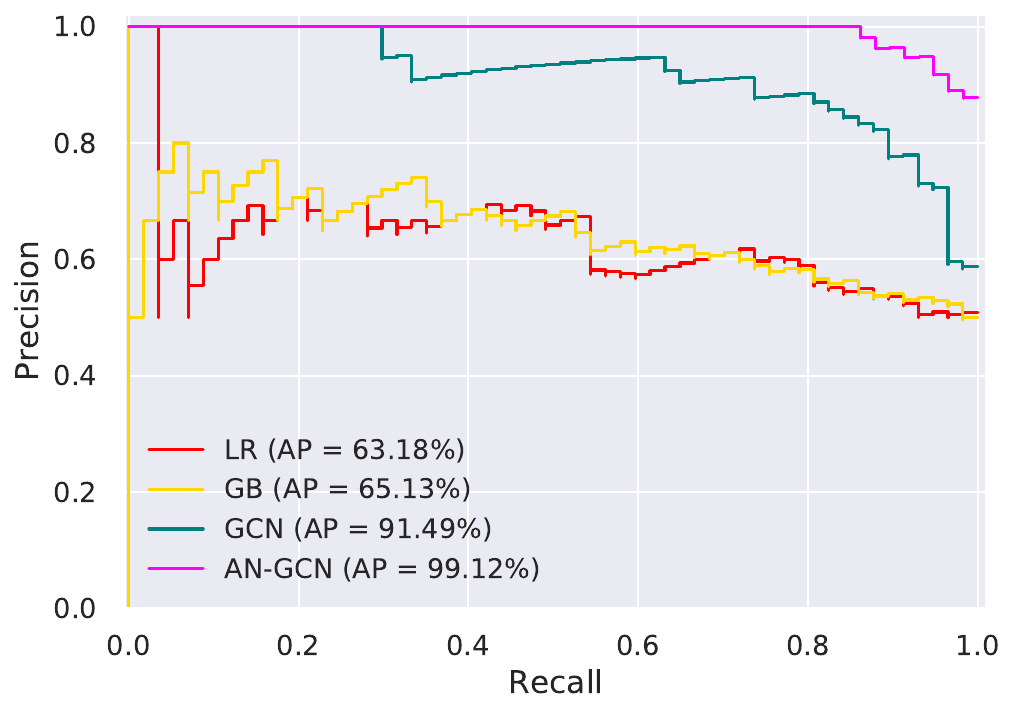}\\
\includegraphics[scale=.525]{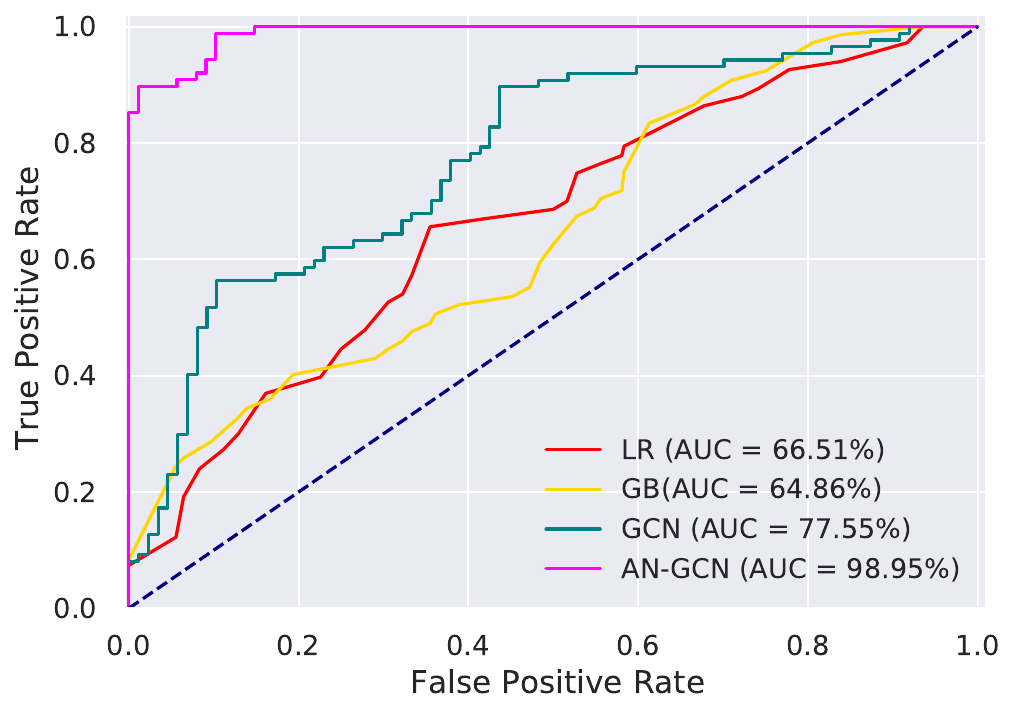}
\caption{Precision-Recall and ROC curves of our model and baseline methods on the ADNI dataset. Average precision (AP) and AUC values are enclosed in parentheses.}
\label{Fig:ADNI_recall_precision}
\end{figure}

Figures~\ref{Fig:ABIDE_recall_precision} and~\ref{Fig:ADNI_recall_precision} show that our model yields the best performance compared to the baselines on both ABIDE and ADNI datasets. As can be seen, the PR (resp. ROC) curve of our model is much closer to the upper right (resp. left) than the corresponding curves for the baselines, indicating the better performance of AN-GCN in disease prediction tasks. In the ROC curves, the diagonal dashed line, which depicts a random algorithm (i.e., random guessing of classes), divides the ROC space. Points above the diagonal represent good classification results (better than random), points below the line poor results (worse than random). Notice that the ROC curves of the logistic regression and gradient boosting are closer to the diagonal line on the ABIDE dataset, indicating poor classification performance.

Overall, our AN-GCN model outperforms GCN and the other baselines significantly and consistently across all datasets, achieving state-of-the-art results in terms of various performance evaluation metrics. In particular, our model improves over the GCN baseline by a big margin. Another key observation is that AN-GCN also outperforms DeepGCN, yielding relative improvements of 31.47\%, 28.80\% and 39.44\% over GCN in terms of accuracy, AUC and F1-score, respectively, on the ABIDE dataset.

\medskip\noindent{\textbf{Model Efficiency.}}\quad The ability of a model to deliver accurate results while minimizing computational resources is crucial for practical applications, and it becomes particularly important in the context of the proposed AN-GCN model for the classification of developmental and brain disorders. For simplicity, we assume the embedding dimensions are the same for all layers, i.e., $F_{\ell}=F$ for all $\ell$, with $F \ll N$. By employing the AN-GCN propagation rule described in Eq.~\eqref{Eq:K_layers_aggregation_model}, we can show that our AN-GCN model has the same memory and time complexity as GCN. In fact, to evaluate the memory complexity, we can observe that an $L$-layer AN-GCN requires $\mathcal{O}(LNF+LF^2)$ in memory, where $\mathcal{O}(LNF)$ is for storing all embeddings and $\mathcal{O}(LF^2)$ is for storing all layer-wise weight matrices. For time complexity, the right-hand side term of the AN-GCN propagation rule has complexity $\mathcal{O}(\vert\mathcal{E}\vert F + NF^2)$, where $\vert\mathcal{E}\vert$ denotes the number of graph edges. Indeed, multiplying the adjacency matrix with an embedding costs $\mathcal{O}(\vert\mathcal{E}\vert F)$ in time, while multiplying an embedding with a weight matrix costs $\mathcal{O}(NF^2)$. Note that the aggregated feature diffusion rule, as described by Eq.~\eqref{Eq:AFD}, also costs $\mathcal{O}(\vert\mathcal{E}\vert F)$ due to element-wise matrix multiplication. Hence, an $L$-layer AN-GCN requires $\mathcal{O}(L\vert\mathcal{E}\vert F+LNF^2)$. Therefore, the proposed AN-GCN propagation rule demonstrates that our model has the same memory and time complexity as the standard GCN. This means that the computational requirements of our AN-GCN model are comparable to those of GCN-based baselines, ensuring efficient implementation and scalability. By maintaining similar complexities, our model offers a practical advantage as it also utilizes skip connections, identity mapping, and an aggregation mechanism to enhance the model performance without introducing significant computational overhead.

\subsection{Parameter Sensitivity Analysis}
In order to answer \textbf{RQ2} and \textbf{RQ3}, we analyze the sensitivity of our disease prediction model to the choice of the number of network layers and the batch size. As the number of network layers plays an important role, we first study how the performance changes as a function of the network depth. Then, we study the performance variation for our model with respect to the batch size on both ABIDE and ADNI datasets.

\begin{figure}[!htb]
\centering
\includegraphics[scale=.525]{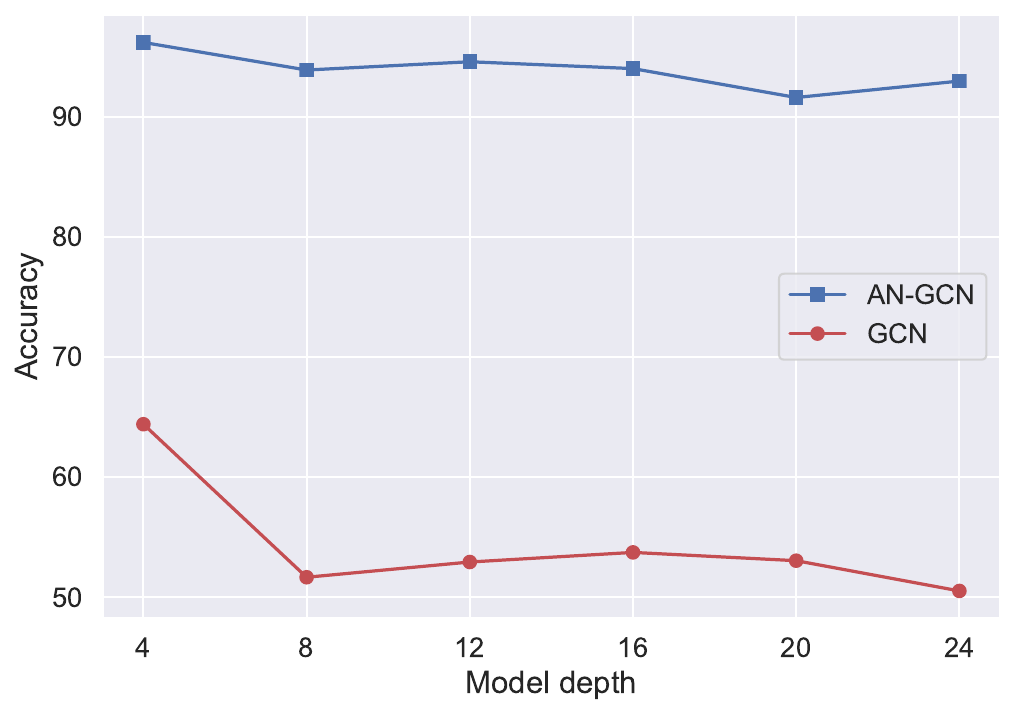}\\[1ex]
\includegraphics[scale=.525]{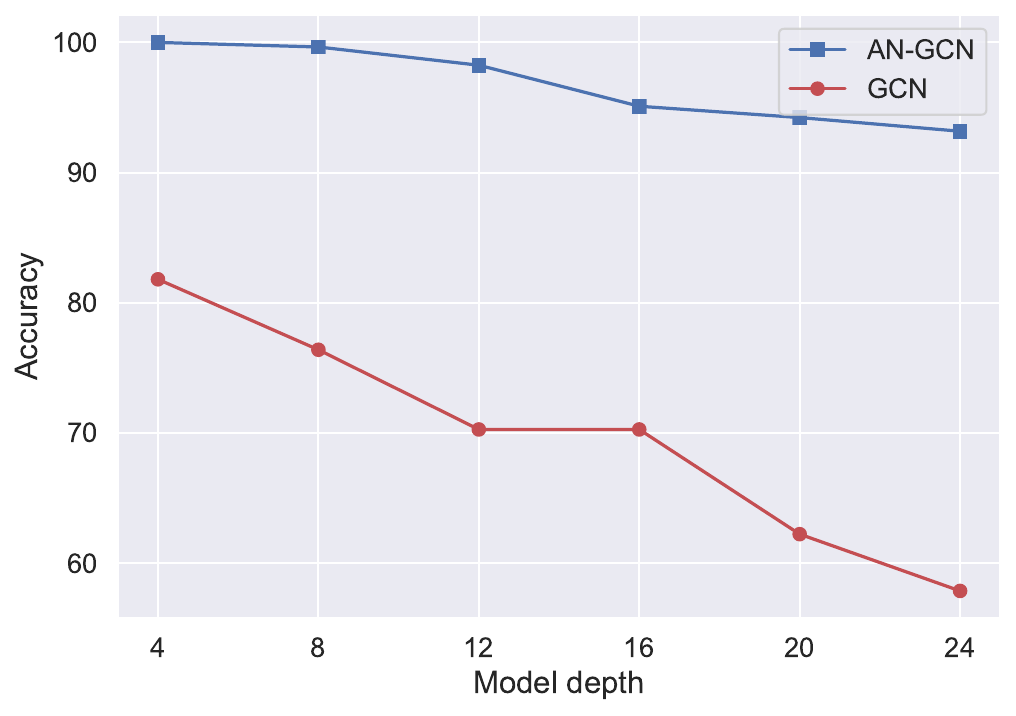}
\caption{Performance comparison of AN-GCN and GCN on the ABIDE (top) and ADNI (bottom) datasets as we increase the number of layers.}
\label{Fig:LayreSize}
\end{figure}

\medskip
\noindent{\textbf{Mitigating the Oversmoothing Problem.}}\quad To evaluate the robustness of our approach to oversmoothing, we study
the performance variation for our multi-layer AN-GCN model on the ABIDE and ADNI datasets with respect to the number of layers. Figure~\ref{Fig:LayreSize} shows how the node classification accuracy changes with the network's depth. As can be seen, the performance of AN-GCN does not significantly degrade compared to GCN when the number of layers increases. Moreover, the performance gap between AN-GCN and GCN becomes quite noticeable when the network's depth rises. Hence, the classification performance of AN-GCN remains relatively stable as we increase the number of layers, demonstrating the robustness of our model against oversmoothing. This is largely due to the fact that the aggregation scheme of the proposed approach leverages residual connections to help alleviate the oversmoothing problem.

\begin{figure}[!htb]
\centering
\includegraphics[scale=.525]{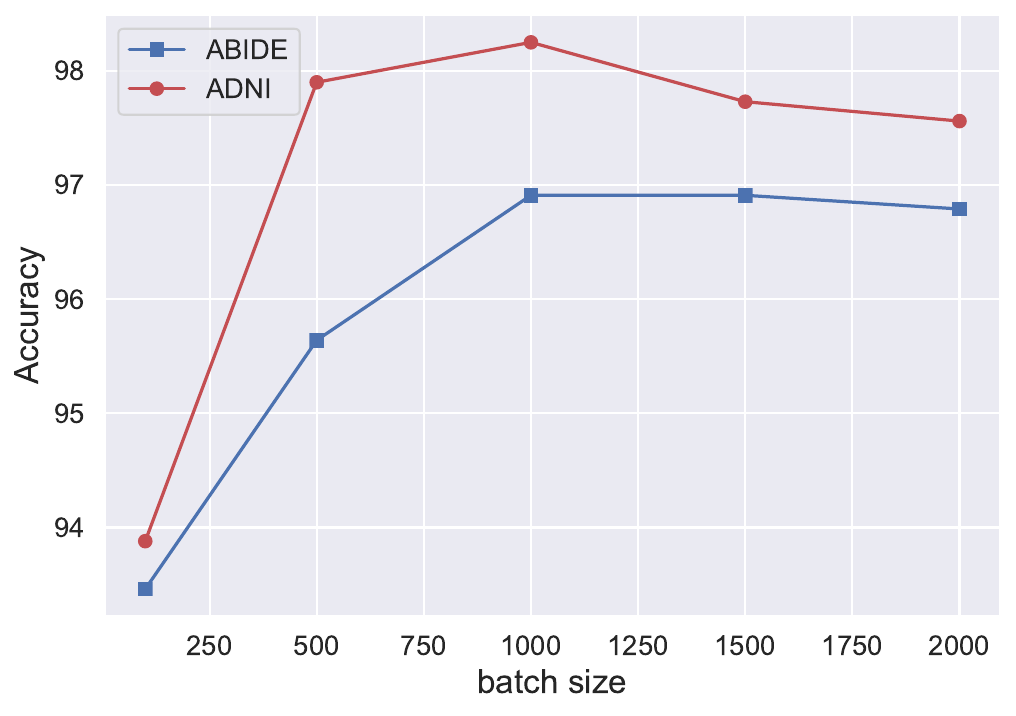}
\caption{Sensitivity analysis of our model to the batch size on the ABIDE and ADNI datasets.}
\label{Fig:BatchSize}
\end{figure}

\medskip
\noindent{\textbf{Effect of Batch Size.}}\quad We test the performance of our model using different values for the batch size on the ABIDE and ADNI datasets. As shown in Figure~\ref{Fig:BatchSize}, the classification accuracy increases rapidly at the beginning (i.e., for smaller batch sizes), reaching the highest value when the batch size is equal to 1000, and then slowly starts to decline on the ABIDE dataset or slows down on the ADNI dataset. This indicates that the batch size also plays an important role. In fact, we can observe that using a large batch size to train our model allows computational speedups from the parallelism of GPUs, but a larger batch size leads to poor generalization. It should also be pointed out that the drawback of using a smaller batch size is that the model is not guaranteed to converge to the global optimum.

\subsection{Limitations}
While our AN-GCN model offers several advantages, it is important to acknowledge two main limitations. First, understanding the underlying features that contribute to the model's predictions can be challenging, especially in the context of complex neurological conditions. While skip connections and identity mapping contribute to improved model performance, they can make the model less interpretable. Interpreting the learned representations and identifying the most relevant features for diagnosis and decision-making may require additional efforts. Second, the effectiveness of the model in real-world clinical settings and diverse patient populations needs to be thoroughly evaluated and validated, albeit it achieves competitive performance on benchmark datasets. In fact, the model's performance may degrade when applied to data from a different domain or population that exhibits significant differences from the training data. Factors such as variations in data collection protocols, demographics, and cultural contexts can contribute to domain shift.

\section{Conclusion}
In this paper, we introduced a graph convolutional aggregation model by learning discriminative node representations from a population graph, consisting of subjects as nodes and edges as connections between subjects, with the goal of predicting the status of each subject (i.e., diseased or healthy control) using imaging and non-imaging features associated to the graph nodes and edges, respectively. The proposed framework leverages skip connections and identity mapping, as well as aggregation in graph sampling in a bid to alleviate the problem of over-smoothing in graph convolutional networks. We demonstrated through extensive experiments that our model outperforms existing graph convolutional based methods for disease prediction on two large benchmark datasets, achieving significant relative improvements in classification accuracy over GCN and other strong baselines. Our AN-GCN model demonstrates an interesting characteristic: it shares the same memory and time complexity as the standard GCN. This similarity in computational requirements ensures that our model can be efficiently implemented and scaled, just like other GCN-based baselines. Notably, our model offers a practical advantage by incorporating skip connections, identity mapping, and an aggregation mechanism to enhance its performance, all without introducing significant computational overhead. This combination of improved performance and comparable complexity makes our model a promising choice for practical applications, where computational efficiency is a crucial consideration. Potential practical implications of our model encompass improved diagnostic accuracy, early detection and intervention, and personalized treatment strategies. Early intervention, for instance, can lead to better treatment outcomes and improved quality of life for individuals affected by developmental and brain disorders. By accurately classifying individuals into diseased or healthy control groups, our model provides valuable insights for personalized treatment planning. The identification of specific patterns or features associated with disease status can help clinicians understand the underlying mechanisms of the disorder and guide treatment decisions. For example, the model may identify imaging or non-imaging features that are highly indicative of disease presence or severity, providing important clues for targeted interventions. For future work, we plan to integrate higher-order graph convolutions into our model with the aim of capturing long-range dependencies between subjects in a population graph. We would also like to investigate the tradeoff introduced by the hyperparameters of the layer-wise propagation rule of our model with the purpose of gaining more theoretical insight. In addition, we intend to apply our model to data relational graphs, where nodes can be connected to each other via multiple relations.

\bibliography{references} 
\bibliographystyle{vancouver}

\end{document}